\documentclass[11pt,a4paper]{article}

\usepackage[hyperref]{emnlp2020}
\usepackage{times}
\usepackage{latexsym}

\usepackage{graphicx}
\usepackage{amsmath}
\usepackage{amsthm}
\usepackage{booktabs}
\usepackage{algorithm}
\usepackage{array}
\usepackage{subfigure}
\usepackage{multirow}
\usepackage{threeparttable}
\usepackage{tabularx}
\usepackage{booktabs}
\usepackage{colortbl}
\usepackage{xcolor}
\usepackage{stfloats}
\usepackage[noend]{algpseudocode}
\usepackage{mathtools}
\usepackage{makecell}

\usepackage{CJKutf8}

\newenvironment{itemize*}%
 {\begin{itemize}%
  \setlength{\itemsep}{0pt}%
  \setlength{\parskip}{0pt}}%
 {\end{itemize}}
 
\usepackage{microtype}

\aclfinalcopy % Uncomment this line for the final submission
\title{Generating Adversarial Examples in Chinese Texts Using Sentence-Pieces }

 \author{
 Linyang Li, Yunfan Shao\thanks{\ \ Equal Contribution.} , Demin Song , Xipeng Qiu\thanks{\ \  Corresponding author.}, Xuanjing Huang  \\
  Shanghai Key Laboratory of Intelligent Information Processing, Fudan University \\
  School of Computer Science, Fudan University \\
  \texttt{\{linyangli19, yfshao19, dmsong20, xpqiu, xjhuang\}@fudan.edu.cn}
  }

\date{}

\begin{document}
\maketitle

\begin{abstract}

Adversarial attacks in texts are mostly substitution-based methods that replace words or characters in the original texts to achieve success attacks.
Recent methods use pre-trained language models as the substitutes generator. 
While in Chinese, such methods are not applicable since words in Chinese require segmentations first.
In this paper, we propose a pre-train language model as the substitutes generator using sentence-pieces to craft adversarial examples in Chinese.
The substitutions in the generated adversarial examples are not characters or words but \textit{'pieces'}, which are more natural to Chinese readers.
Experiments results show that the generated adversarial samples can mislead strong target models and remain fluent and semantically preserved.

\end{abstract}

\section{Introduction}
Adversarial attacks \cite{goodfellow2014explaining,kurakin2016adversarial,chakraborty2018adversarial} are firstly introduced in the computer vision fields.
Neural networks are vulnerable to adversarial samples with small perturbations based on gradients to the original inputs.

In the natural language processing fields, perturbations cannot be applied directly, so most methods involve strategies such as replacing, insertion or deletion \cite{ebrahimi2017hotflip,Alzantot,jia2017adversarial,jin2019textfooler}.
Most replacing strategies incorporate a synonym dictionary \cite{dong-etal-2010-hownet,wordnet} or use word embeddings \cite{pennington2014glove,mrksic:2016:naacl, jin2019textfooler}.

Recently, incorporating pre-trained language models such as BERT \cite{bert} as a substitute generator is introduced in the adversarial example generation of texts \cite{li2020bert,garg2020bae}.
Such a process brought adversarial example generation to a higher level: using well-learned models to generate adversarial samples instead of specific rules such as entity checking or grammar checking \cite{jin2019textfooler}.

However, in Chinese language, crafting adversarial examples are more challenging:
Chinese does not contain explicit whitespace between words, so the boundaries between characters and words are vague.
% Replacing words in Chinese requires segmentations first, so the generated adversarial samples are rather rigid since a pre-processed segmentation would make adversarial attacks less successful.
Therefore, current Chinese pre-trained language models are character-based, yet replacing certain characters in Chinese cannot maintain fluency and semantics.

\begin{figure}[]
\centering
\includegraphics[width=1.0\linewidth]{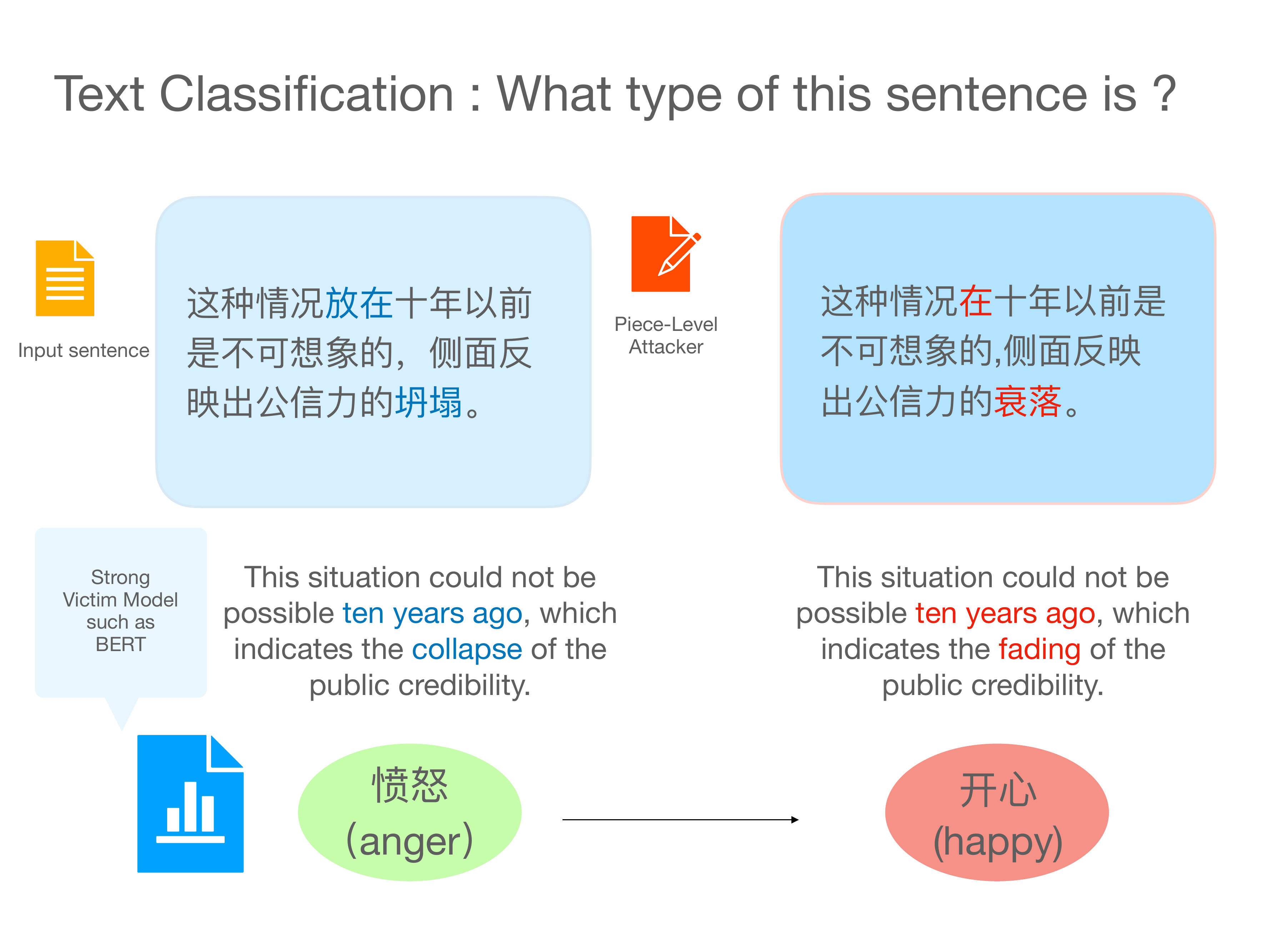}
\centering
\caption{Example of piece-level Adversary}
\label{fig:example}
\end{figure}

Therefore, in this paper, we propose the idea of incorporating sentence-pieces in the Chinese tokenization in pre-trained language models, and use such a language model as a sentence-piece level substitution generator to craft high-quality adversarial examples in Chinese.

At first, we use sentence-piece tokenizer \cite{kudo-richardson-2018-sentencepiece} to create a piece-level vocabulary based on the large-scale corpus collected online \cite{bright_xu_2019_3402023}. 
After sentence-piece tokenization, we use this piece-level vocabulary to pre-train a masked language model as the substitution generator following the standard pre-training process of BERT.
Then we use the pre-trained substitution generator to craft piece-level adversarial samples in Chinese.
As seen in Fig\ref{fig:example}, we can generate character-level, word-level and phrase-level substitutions in Chinese.

We use the trained Chinese Substitution Generator to attack broadly used Chinese text classification tasks such as Sogou, Iflytek, Weibo and Law34 datasets.
The experiments show that the generated adversarial samples successfully mislead target models and reserve the semantic information and fluency.

To summarize the key contribution of this paper:
we use sentence-piece based tokenization to train a masked language model that can generate piece-level substitutions for Chinese adversarial examples.
To our knowledge, we are the first to craft adversarial samples in Chinese that can generate multi-level substitutions.

\section{Backgrounds}

\subsection{Adversarial Attacks in NLP}
In the NLP field, adversarial attacks face a major challenge: people cannot apply gradients on the embedding space to craft adversarial samples, which is widely explored in the CV field \cite{goodfellow2014explaining, chakraborty2018adversarial}.
So replacing characters \cite{ebrahimi2017hotflip}, words \cite{Alzantot,jin2019textfooler,ren2019generating,papernot2016crafting} or paraphrasing sentences \cite{jia2017adversarial} are the main streams of generating adversarial samples in texts.

The formulation of attacking NLP models is that, given inputs $X = [w_0, \cdots, w_i, \cdots]$, where $w_i$ is the target character/word, we have a candidate list $S(w_i) = [s_{i0}, \cdots, s_{ij} ]$, which is the potential substitutes of $w_i$.
The goal is to find an adversarial sample $X^{'} = [w_0, \cdots, s_{id}, \cdots ]$ where $ argmax(F(X)) {\neq} argmax(F(X^{'}))$.
In most cases, we assume that we know the output score of the target model, which is a black-box scenario, while in white-box attacks we know the model architecture and therefore the gradients over the inputs.

\citet{jin2019textfooler} uses word embeddings to craft candidate list $S$ to find substitutions while \citet{zang2020word} uses knowledge extracted from WordNet \cite{wordnet}.
Recently \citet{li2020bert,garg2020bae,shi2020paraphrase} uses pre-trained models to find similar tokens as the candidate list.

\subsection{Masked Language Models}

Pre-trained models exemplified by BERT \cite{bert} introduces a masked-language model task to predict the masked tokens.
These models \cite{bert} use BPE-based tokenizations to avoid unknown words in English.
With massive calculation, these pre-trained models have revolutionized NLP tasks.

In Chinese texts generation, a whole-word-mask strategy is introduced to predict the entire word properly \cite{cui2019pre}, while the tokenization is still character-level.
Therefore, such a model cannot be directly applied to generate multiple candidates as the substitution list \cite{li2020bert,garg2020bae} in adversarial attacks.

\subsection{Sentence-Piece Tokenization}

Sentence-piece tokenization \cite{kudo-richardson-2018-sentencepiece} is derived from byte-pair encoding \cite{sennrich-etal-2016-neural} and unigram language model \cite{kudo2018subword} with the extension of direct training from raw sentences.
Diffrent from BPE, sentence-piece is directly trained from raw texts so no word segmentation is needed.
Therefore, we could create a vocabulary based on sentence-piece based tokenization in Chinese.

\section{Method}

In this section, we introduce mainly two parts: (1) how we train a \textit{piece-level} masked language model as a Chinese substitution generator; (2) how we use such a model to generate adversarial samples in Chinese.

\subsection{Training the Chinese Substitution Generator}

To train a Chinese Substitution Generator, we follow the standard protocol of pre-training masked language models.
We simply replace the vocabulary with Chinese sentence-pieces in order to generate proper substitutions.

There are only a few thousands of commonly used Chinese characters, we extend the vocabulary size to 60 thousand pieces, based on the training data collected online \cite{bright_xu_2019_3402023}.

The statistics shown in Table \ref{tab:vocab-size} indicates that there are around 9\% (5400) of pieces in the vocabulary are single characters. 
Most pieces are bi-char words or tri-char words.
There are also a considerable amount of pieces with more than 3 characters, which are usually phrases or continuous entities.

\begin{CJK*}{UTF8}{gbsn}
\begin{table}[]
    \centering
    \scriptsize
    \begin{tabular}{c|c|c|c}
        Char & Num(\%) & Examples \\
        1 & 9 \% & 是(is),在(in),前(before)\\ 
        2 & 44 \% & 但是(but),这个(this),生活(life)\\
        3 & 24 \% & 自己的(belong to me),是不是(is or not)\\
        4 & 17 \% & 这个问题(this problem)\\
        5+ & 4 \% & 证券投资基金(Securities Investment Funds)\\
    \end{tabular}
    \caption{Statistics of the vocabulary}
    \label{tab:vocab-size}
\end{table}
\end{CJK*}

The pre-trained model is an architecture the same as BERT-base, with 12 layers and hidden size set to 768.
We use LAMB optimizer \cite{lamb} to pre-train our model on NVIDIA 3090 GPUs using fairseq toolkit \cite{ott2019fairseq}.

\subsection{Generating Adversarial Examples}

In crafting adversarial examples, we apply a two-step algorithm which is widely used in crafting substitution-based adversarial examples \cite{jin2019textfooler,li2020bert}:
(1) first we find the most vulnerable pieces by iteratively ranking the piece importance of the original input sentence. 
We follow \citet{jin2019textfooler,li2020bert} to measure the piece importance by masking tokens iteratively and calculate the output scores:
\begin{align}
    I_{w_i} = o_y(S) - o_y(S_{\backslash w_i}) , \label{eq:importance}
\end{align}
where $S_{\backslash w_i} = [w_0, \cdots, w_{i-1}, [\texttt{MASK}], w_{i+1}, \cdots]$ is the sentence after replacing $w_i$ with $[\texttt{MASK}]$ and $o_y()$ is the output score after the softmax function in the final classification layer.

Then we replace pieces with the candidates predicted by the Chinese Substitution Generator.

Following \citet{li2020bert}, we do not mask the original pieces so that the semantic information is preserved. 
We use the top-$K$ predictions of the given pieces in the masked language model as the substitution candidates.
Since the tokenization process is piece-level, the substitution candidates are flexible: candidates contain both single characters, phrases that expand the meaning of the original piece, words that have similar meanings with the original piece.

\begin{CJK*}{UTF8}{gbsn}
As seen in Table \ref{tab:topk-pred}, we have different types of substitutions of a single piece: we can replace Chinese word '今天(today)' with a synonym '今日(today)', or we can find an expanded word '今天的(today's)'; also, we can replace it with a single character '今(now/', which is a shorten version of 'today'. 
With the piece-level substitution generator, the generated examples could be very diversified.

\end{CJK*}

\begin{CJK*}{UTF8}{gbsn}

\begin{table}[]
    \centering
    \small
    \begin{tabular}{c|cccccccc}
        pieces & candidate types & candidates list \\
         \multirow{3}*{今天} & words & 今日 \\
        & phrase & 今天的\\
        & character &  今 \\
        \midrule

    \end{tabular}
    \caption{eaxmple of top-k predictions}
    \label{tab:topk-pred}
\end{table}

\end{CJK*}

After getting candidate list, we replace pieces by the order of the ranked word importance.
In practice, the candidate list size $K$ is 12 in all datasets.
We find the most harmful piece in the candidate list as the perturbation of the current piece, if the model cannot correctly predict the classification of the sample, we return the generated adversarial sample.
Otherwise, we continue to find another piece to replace until we find a proper adversarial example.

\section{Experiments}

\subsection{Datasets and Victim Models}

We use several popular Chinese text classification tasks as our attacking datasets \footnote{https://github.com/LinyangLee/CN-TC-datasets}\footnote{https://github.com/CLUEbenchmark/CLUE}:

Sogou : the Sogou dataset is a 12-class sentence-genre task.
    
IflyTek: the Iflytek dataset is part of the Chinese GLUE benchmark, and it is a 119-class sentence genre task. 

Weibo : the Weibo dataset is a sentiment classification task containing 8 emotions.

Law34 : the Law34 dataset is a 34-class task predicting the court decision type of the given texts.

We fine-tune the standard BERT-base-chinese model as our victim models for the corresponding tasks \footnote{https://github.com/google-research/bert} with huggingface transformers \cite{wolf-etal-2020-transformers}.

We randomly select 200 examples from the development set in each dataset and craft their corresponding adversarial examples.

\subsection{Evaluation of the Generated Examples}

The major metric is the attack success rate, which is the percentage of successful attacks in the dataset.
The second metric is the change rate of the generated adversarial samples.
We intuitively believe that fewer changes could result in a less semantic shift.

Further, we run a human evaluation to measure the quality of the generated adversarial samples.
We mix original samples and generated adversaries and ask human judges to predict the label and the fluency of the examples.
% human-eval ? 都需要哪些 ？
Since there could be too much labels for human judges, we ask human judges to predict whether the given texts are the correct label or not.
We also ask them to score the fluency ranging from 1-5.

\subsection{Baselines}

We setup a strong baseline to compare with our method:

Char-Replace: We incorporate the original Chinese BERT which is character-level to generate adversarial samples by replacing characters.
The hyper-parameters are the same as our piece-level model.

Word-Replace: We use a 50-dimension word-embedding collected by \citet{zhang2018chinese} and use cosine-similarity to find candidate list as done by \citet{jin2019textfooler}.
Since it is a word-level attack, we use Jieba \footnote{https://github.com/fxsjy/jieba} tokenization tool to tokenize the sequence. 
We use a threshold of the cosine-similarity to constrain the quality of the candidates.
We use around 60K most-frequent words in the word-embedding.

\begin{table}[]\setlength{\tabcolsep}{2pt}
    
    \centering
    
    \scriptsize
    \begin{tabular}{c|c|cclcccc}
    \toprule
    
    \multirow{2}*{\textbf{Dataset}}  & \multirow{2}*{Method} & \multirow{2}*{Ori Acc} & \multirow{2}*{Atk Acc} & \multirow{2}*{Ptb\%}   & \multicolumn{2}{c}{Human}\\
    &&&&&  Consist \% & Fluency  \\
    \midrule
    \multirow{3}*{IflyTek} & Char-R & \multirow{3}*{60.0} & 9.0 & 2.7 (c) & 90 & 4.2 & \\    
    & Word-R \textbf{} &  & 10.0 & 2.5 (w) & 92 & 4.3 \\ 
    & Pce-R(\textbf{ours}) &  & 11.0 & 2.7 (pce) & 94 & 4.5 \\ 
    \midrule
    
    \multirow{3}*{Weibo} & Char-R & \multirow{3}*{92.0} & 27.0 & 9.1 (c) & 88 & 4.0 \\    
    & Word-R \textbf{} &  & 30.0 & 8.7 (w) & 89 & 4.0\\ 
    & Pce-R(\textbf{ours}) &  & 35.0 & 10.2 (pce)  & 90 & 4.2 \\ 
    \midrule
    
    \multirow{3}*{Sogou} & Char-R & \multirow{3}*{93.5} & 15.0 & 3.2 (c) & 90 & 3.8\\    
    & Word-R \textbf{} &  & 16.0 & 3.0 (w) & 91 & 3.9 \\ 
    & Pce-R(\textbf{ours}) &  & 18.5 & 4.4 (pce)  & 93 & 4.2\\ 
    \midrule
    
    \multirow{3}*{Law34} & Char-R & \multirow{3}*{93.0} & 10.0 & 2.8 (c) & 94 & 4.5 \\    
    & Word-R \textbf{} &  & 11.0 & 2.7 (w) & 95 & 4.6\\ 
    & Pce-R(\textbf{ours}) &  & 12.0 & 4.2 (pce) & 97 & 4.8 \\ 
    \bottomrule

    \end{tabular}
    \caption{Main Results of Generated  Examples \\}
    \label{tab:main}
\end{table}

\subsection{Main Results of Generated  Examples}

As seen in Table \ref{tab:main}, the generated adversarial samples successfully mislead the strong fine-tuned BERT-chinese models.
Also, human judges give a high accuracy predicting whether the classification is correct, also give a high fluency score.

Compared with the character-level and word-level method, the piece-level adversarial samples can achieve similar attacking results and maintain a high fluency score.
The adversarial samples should be both harmful to the target models and semantically fluent, therefore the piece-level adversarial samples generated are better adversarial samples though the success rate is lower than the character-replace and word-replace methods.
The success rate could be higher when expanding the size of the candidate list according to \citet{Morris2020ReevaluatingAE} therefore what matters most is the quality of the adversarial samples.

As seen in the appendix, the generated piece-level adversarial samples could make successful attacks from different aspects and remain fluent while the character-level adversarial samples are harder to comprehend.
We can replace tokens with similar meaning substitutes, and we can also replace them with irrelevant but fluent and label-preserved substitutes.

\subsection{Trade-off between Candidate List Size and Success Rate}

A larger candidate size would result in easier attacks.
Therefore we apply a trade-off curve showing that we can achieve a significantly higher attack success rate when we use a large candidate size.
But it is intuitive that larger candidate size may also significantly harm the quality of the generated adversarial samples in both semantics and fluency.

Therefore we believe that \textbf{success rate is not the most important}, since maintaining the fluency and semantic is one major concern in crafting adversarial samples.
While in our experiment, piece-level candidates are generally more natural to human judges.

\begin{figure}[]
\centering
\includegraphics[width=0.9\linewidth]{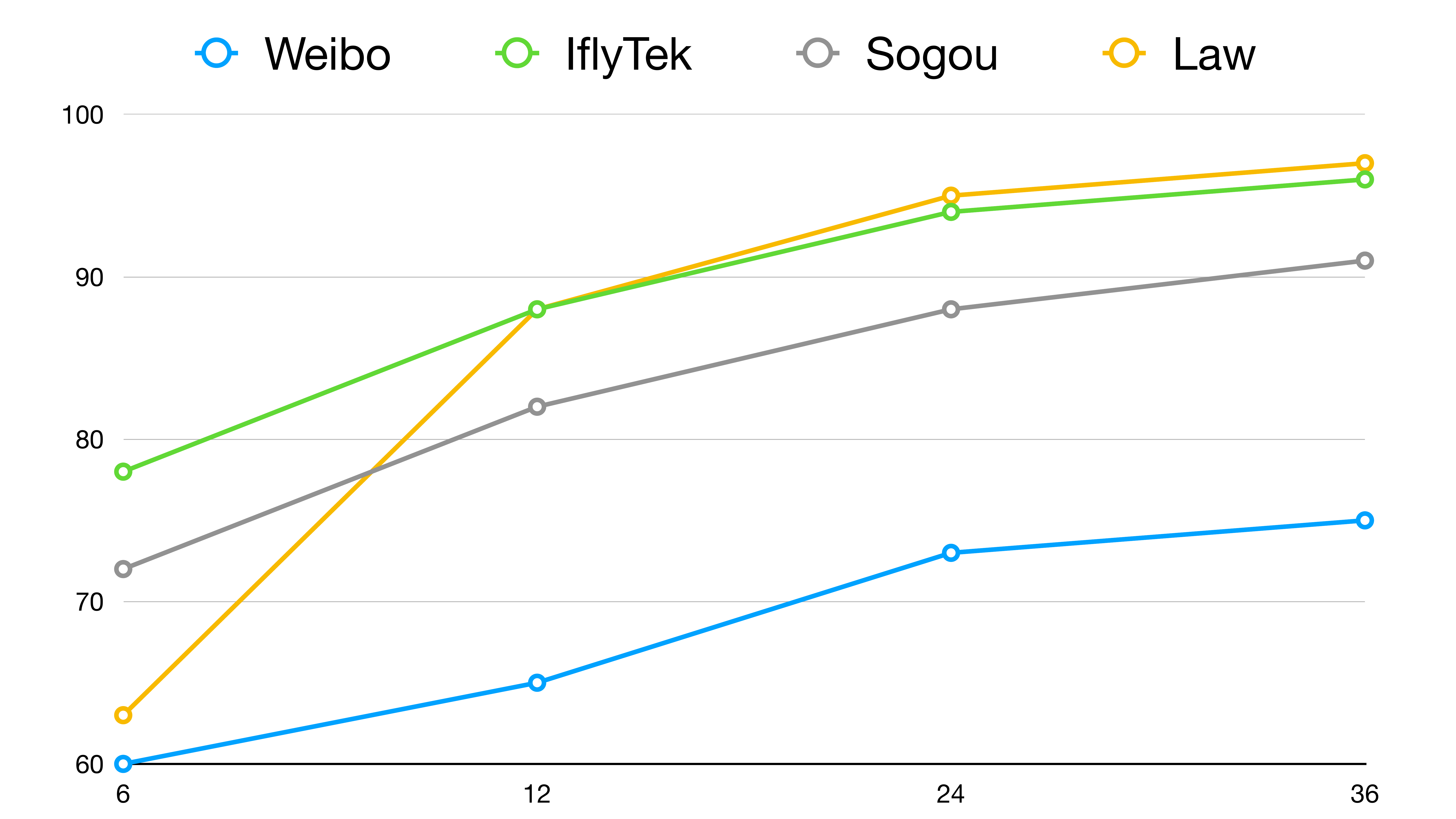}
\centering
\caption{Trade-Off curve between $K$ and Success}
\label{fig:tradeoffcurve}
\end{figure}

\section{Conclusion}
In this paper, we propose a piece-level adversarial sample generation strategy for Chinese texts, which can fill in the blank of text adversarial sample generation for languages other than English.

\newpage

\bibliography{emnlp2020}
\bibliographystyle{acl_natbib}

\appendix

\section{Appendices}
\label{sec:appendix}

Here we provide some of the generated adversarial samples.

As seen in Table \ref{tab:samples}, piece-level substitutions could have various types of strategies to craft adversaries:

\begin{itemize}
    \item Synonym Replacing: Using synonyms or similar pieces as substitutions.
    
    \item Rewrite: Paraphrasing the given pieces.
    
    \item Expansion: Expand the given pieces.
    
\end{itemize}

\begin{CJK*}{UTF8}{gbsn}
\newcolumntype{L}[1]{>{\raggedright\let\newline\\\arraybackslash\hspace{0pt}}m{#1}}
\begin{table*}[ht]\setlength{\tabcolsep}{6pt}\scriptsize
    
    \begin{tabular}{lllllll}
        \toprule 
        \textbf{Dataset}&&&\text{Label} \\
        \midrule

        \multirow{8}{*}{\bfseries Law} &
        {Ori} &
        \makecell[l]{被告人黎某甲伙同绰号“杨某”等人去到该房内对严某乙进行殴打, 之后黎某甲等人又拿着\\
        铁棍、木棍等追赶乘坐小汽车离开的严某乙至苍梧大道广场罗马柱对出街道处，将小汽车\\
        的挡风玻璃等物品砸烂，并将严某乙\textcolor[rgb]{0.00,0.07,1.00}{打伤}。经法医鉴定严某乙所受损伤构成轻伤。\\
        (The defendant Li along with his associates, 'Yang' et al. went to the room to beat Yan. \\
        Then they chased after Yan who was leaving in a car with iron rods and wooden sticks. \\
        On the street facing the Roman pillar of Cangwu Avenue Square, they smashed the windshield \\
        of the car and \textcolor[rgb]{0.00,0.07,1.00}{injured} Yang. According to the forensic examination, \\
        Yan's injury constituted a minor injury.)\\} 
        &\textcolor[rgb]{0.00,0.07,1.00}{\makecell[l]{故意伤害\\Intentional injury}}\\

        \cmidrule{2-4} &{Adv} &
        \makecell[l]{被告人黎某甲伙同绰号“杨某”等人去到该房内对严某乙进行殴打, 之后黎某甲等人又拿着\\
        铁棍、木棍等追赶乘坐小汽车离开的严某乙至苍梧大道广场罗马柱对出街道处，将小汽车\\
        的挡风玻璃等物品砸烂，并将严某乙\textcolor{red}{重伤}。经法医鉴定严某乙所受损伤构成轻伤。\\
        (The defendant Li along with his associates, 'Yang' et al. went to the room to beat Yan. \\
        Then they chased after Yan who was leaving in a car with iron rods and wooden sticks. \\
        On the street facing the Roman pillar of Cangwu Avenue Square, they smashed the windshield \\
        of the car and \textcolor{red}{seriously injured} Yang. According to the forensic examination, \\
        Yan's injury constituted a minor injury.)}
        &\textcolor{red}{\makecell[l]{酒驾\\Drunk driving}}\\
        
        \midrule

        \multirow{4}{*}{\bfseries Weibo} &
        {Ori} &
        \makecell[l]{你\textcolor[rgb]{0.00,0.07,1.00}{竟然还}给我\textcolor[rgb]{0.00,0.07,1.00}{唱}《夜上海》《给我一个吻》啊啊啊啊啊啊啊啊啊啊啊啊！！！\\
        You \textcolor[rgb]{0.00,0.07,1.00}{even sang} "Night Shanghai" and "Give Me a Kiss" to me. OMG!} 
        &\textcolor[rgb]{0.00,0.07,1.00}{开心(Happy)}\\

        \cmidrule{2-4} &{Adv} &
        \makecell[l]{你\textcolor{red}{还可以}给我\textcolor{red}{唱的}《夜上海》《给我一个吻》啊啊啊啊啊啊啊啊啊啊啊啊！！！\\
        You \textcolor{red}{could also sang} "Night Shanghai" and "Give Me a Kiss" to me. OMG!}
        &\textcolor{red}{嫌弃(Disgust)}\\

        \midrule

        \multirow{7}{*}{\bfseries Iflytek} &
        {Ori} &
        \makecell[l]{英雄联盟比赛视频全球英雄联盟职业比赛赛程全球职业选手比赛数据及视频职业选手\\
        \textcolor[rgb]{0.00,0.07,1.00}{符文},
        每局比赛所有选手数据查询优质\textcolor[rgb]{0.00,0.07,1.00}{咨询}内容.玩加电竞是一款为游戏玩家而生的APP，\\
        是将比赛赛程，数据，战绩查询、新闻资讯、玩家社区、小组等结合在一起的APP\\
        (League of Legends competition video; League of Legends professional competition schedule; \\ 
        global professional players competition data, video and \textcolor[rgb]{0.00,0.07,1.00}{runes} of Professional Players;\\
        You can find data of every game of all professional players and high-quality \textcolor[rgb]{0.00,0.07,1.00}{information}. \\
        WanJia Gaming is an APP born for game players.)} 
        &\textcolor[rgb]{0.00,0.07,1.00}{MOBA}\\

        \cmidrule{2-4} &{Adv} &
        \makecell[l]{英雄联盟比赛视频全球英雄联盟职业比赛赛程全球职业选手比赛数据及视频职业选手\\
        \textcolor{red}{直播},
        每局比赛所有选手数据查询优质\textcolor{red}{直播}内容.玩加电竞是一款为游戏玩家而生的APP，\\
        是将比赛赛程，数据，战绩查询、新闻资讯、玩家社区、小组等结合在一起的APP\\
        (League of Legends competition video; League of Legends professional competition schedule; \\ 
        global professional players competition data, video and \textcolor{red}{live-streams} of Professional Players;\\
        You can find data of every game of all professional players and high-quality \textcolor{red}{live-streams}. \\
        WanJia Gaming is an APP born for game players.)}
        &\textcolor{red}{\makecell[l]{视频\\Video}}\\

        \midrule

    \end{tabular}
    \caption{Examples of Generated Adversaries
    }

    \label{tab:samples}
\end{table*}
\end{CJK*}

\end{document}